\documentclass[runningheads]{llncs}

\usepackage[T1]{fontenc}
\usepackage{newtxtext,newtxmath} 
\usepackage{amsmath}
\usepackage{graphicx}
\usepackage{booktabs}
\usepackage{multirow}
\usepackage{bm}
\usepackage{bbm}
\usepackage{lineno}

\usepackage[dvipsnames,table]{xcolor}
\usepackage{colortbl}
\usepackage{pifont}
\usepackage{placeins}
\usepackage{enumitem}
\setlist[itemize]{label=\textbullet}

\usepackage[labelfont=bf,labelsep=period,font=small]{caption} 
\usepackage{subcaption}
\usepackage[normalem]{ulem}
\usepackage{array}       
\usepackage{makecell}
\usepackage{microtype}

\renewcommand{\small}{\fontsize{9pt}{11pt}\selectfont}

\newcommand{\wout}{\textit{w/o}}

\newcolumntype{C}[1]{>{\centering\arraybackslash}p{#1}}

\definecolor{lavender}{RGB}{230,230,250}
\definecolor{lightblue}{RGB}{173, 216, 230}

\usepackage[colorlinks=true, citecolor=purple]{hyperref}

\begin{document}

\title{\texttt{CGSTA}: Cross-Scale Graph Contrast with Stability-Aware Alignment for Multivariate Time-Series Anomaly Detection}

\titlerunning{\texttt{CGSTA}: Cross-Scale Graph Contrast with Stability-Aware Alignment}

\author{
Zhongpeng Qi\inst{1}\orcidID{0009-0009-2063-4338} \and
Jun Zhang\inst{1}\thanks{Corresponding author; $^\ddag$ Last author.}\orcidID{0009-0001-4087-6336} \and
Wei~Li\inst{2}\orcidID{0000-0003-0998-5435} \and
Zhuoxuan Liang\inst{2}$^\ddag$\orcidID{0009-0008-7141-5963}
}

\authorrunning{Z. Qi et al.}
\institute{
Dalian Maritime University, Dalian, China\\
\email{qzp@dlmu.edu.cn, zhangjun@dlmu.edu.cn}
\and
Harbin Engineering University, Harbin, China\\
\email{wei.li@hrbeu.edu.cn, zz.liang@hrbeu.edu.cn}
}

\maketitle

\begin{abstract}
\small
Multivariate time-series anomaly detection is essential for reliable industrial control, telemetry, and service monitoring. However, the evolving inter-variable dependencies and inevitable noise render it challenging. Existing methods often use single-scale graphs or instance-level contrast. Moreover, learned dynamic graphs can overfit noise without a stable anchor, causing false alarms or misses. To address these challenges, we propose the \texttt{CGSTA} framework with two key innovations. 
First, Dynamic Layered Graph Construction (DLGC) forms local, regional, and global views of variable relations for each sliding window; rather than contrasting whole windows, Contrastive Discrimination across Scales (CDS) contrasts graph representations within each view and aligns the same window across views to make learning structure-aware. Second, Stability-Aware Alignment (SAA) maintains a per-scale stable reference learned from normal data and guides the current window’s fast-changing graphs toward it to suppress noise. 
We fuse the multi-scale and temporal features and use a conditional density estimator to produce per-time-step anomaly scores. Across four benchmarks, \texttt{CGSTA} delivers optimal performance on PSM and WADI, and is comparable to the baseline methods on SWaT and SMAP.
\end{abstract}

\section{Introduction}
\textbf{Background.} Anomaly detection in multivariate time series (MTS) is crucial for ensuring safety and reliability across domains such as industrial control systems—e.g., secure water treatment (SWaT) and water distribution (WADI) testbeds \cite{goh2016swat,adepu2017wadi}—space telemetry (NASA SMAP) \cite{hundman2018detecting}, and large-scale online service KPIs (PSM) \cite{abdulaal2021practical}. 
These systems generate high-dimensional temporal streams with complex and evolving inter-variable dependencies; detecting anomalies remains challenging due to non-stationarity, transient noise, and correlation shifts across variables.

\textbf{Prior Studies.} 
Existing approaches for multivariate time-series anomaly detection (MTSAD) mainly fall into three categories: sequence modeling , graph-based and  contrastive learning-based approaches. 
\textit{Sequence modeling} methods, such as LSTM \cite{hundman2018detecting} and autoencoder-based models \cite{xu2024couta}, capture temporal dependencies through reconstruction or prediction, but often ignore variable-wise structural relations, limiting their ability to detect cross-variable anomalies. 
\textit{Graph-based} approaches ~\cite{deng2021graph,zheng2023correlation,xu2022anomaly-Transformer,liu2024multivariate}
explicitly model inter-variable correlations with GNNs or attention.
While effective at capturing spatio-temporal patterns, most adopt a single-scale variable graph, which makes them sensitive to transient noise and unable to capture multi-scale dependencies.
Recent progress in \textit{contrastive learning} \cite{oord2018cpc,tang2024perturbation} has improved representation discriminability. However, the prevalent instance-level paradigm—contrasting two augmented views of the same sliding window against views of other windows—often overlooks inter-variable structure \cite{liu2024timesurl,luo2023infots}.
In summary, while above methods have made significant strides in MTSAD, they face two main challenges:
\begin{itemize}[leftmargin=12pt,itemsep=1pt,topsep=2pt]
\item \textbf{Challenge 1: Single-scale view and instance-level contrast.}
Single-scale variable graphs rely on a flat adjacency structure between variables, which is either static (fixed for the entire sequence) or window-based dynamic (recomputed per sliding window) \cite{xu2022anomaly-Transformer,liu2024multivariate,zhou2023mtgflow}. While these designs effectively capture pairwise correlations, they fail to model hierarchical regularities across different views—such as sensor pairs, functional groups, and system-level couplings \cite{ma2023rethinking,zhang2022grelen}.
 Meanwhile, instance-level contrastive objectives \cite{liu2024timesurl} treat each sliding window as an independent unit (e.g., contrasting two augmented views of the same window against others), lacking explicit supervision on inter-variable or cross-scale structures. This combination makes models easily distracted by short-term perturbations, weakening their sensitivity to anomalies that disrupt structural dependencies or involve cooperative deviations across variables.
These observations motivate hierarchical modeling and cross-scale consistency of discriminative cues.
\item \textbf{Challenge 2: Dynamic graphs are noise-sensitive without a stable structural reference.}
Dynamic graphs update adjacency matrices per sliding window to track time-varying inter-variable dependencies \cite{xu2022anomaly-Transformer,zhou2023mtgflow}. However, they are prone to being disrupted by transient noise (e.g., short-lived sensor fluctuations or process disturbances) and hidden anomalies in training data—even “mostly normal” training splits may contain unlabeled anomalous segments that the model inadvertently fits \cite{zhou2023mtgflow,dai2022GANF}. Without a long-horizon structural anchor to capture stable dependencies, the learned graph structure can drift gradually across windows due to noise or non-stationarity. This drift not only causes false alarms when the model misinterprets noise as structural changes but also leads to missed anomalies that distort intrinsic variable dependencies. To address this, a stability-anchored regularization mechanism is needed.
\end{itemize}

\textbf{Proposed Methods.} We present \texttt{the CGSTA} framework, guided by two principles: capturing multi-scale structural dependencies to expose structure-breaking anomalies that may be hidden at a single scale, and maintaining a long-term stable structural reference to filter short-lived noise and avoid drift. \textbf{To address Challenge~1}, Dynamic Layered Graph Construction (DLGC) provides three complementary views per window—local (direct ties between variables), regional (interactions within data-driven groups) and global (relations among groups)—and encodes them into per-scale embeddings; on top of these, Contrastive Discrimination across Scales (CDS) teaches the model what to separate and what to keep consistent by enforcing separation of normal vs.\ abnormal patterns within each scale while aligning representations of the same window across scales, turning instance-level learning into structure-aware discrimination and reducing the omissions of single-scale modeling. \textbf{To address Challenge~2}, Stability-Aware Alignment (SAA) maintains, at each scale, a slowly updated “stable” graph alongside the fast, window-level graph and gently aligns the dynamic estimate to its stable counterpart during training; the stable branch acts as a long-horizon anchor that filters transient fluctuations and prevents gradual drift of learned dependencies, without requiring labels. Finally, \texttt{CGSTA} fuses the multi-scale embeddings and applies a density head to produce anomaly scores, with the overall objective balancing discriminability (CDS) and temporal stability (SAA).

We summarize our contributions as follows:
\begin{itemize}[leftmargin=12pt,itemsep=1pt,topsep=2pt]
  \item We propose a cross-scale graph learning pipeline that couples Dynamic Layered Graph Construction (DLGC) with Contrastive Discrimination across Scales (CDS): DLGC builds local, regional and global dynamic graphs, and CDS turns them into structure-aware representations by enforcing intra-scale separability and cross-scale coherence.
  \item We propose Stability-Aware Alignment (SAA), an EMA-anchored regularization that maintains a long-horizon structural reference and aligns dynamic graphs to suppress transient noise and distribution drift.
 \item We conduct extensive experiments on public benchmarks, demonstrating that \texttt{CGSTA} is effective and robust across datasets.
\end{itemize}

\section{Related Work}
Existing methods for MTSAD can be broadly categorized into three main groups: sequence modeling-based, graph-based and contrastive learning-based approaches.
\ding{182} \textbf{Sequence modeling-based approaches:} Early methods, such as LSTM \cite{hundman2018detecting} and autoencoders \cite{xu2024couta}, rely on sequence modeling to capture temporal dependencies. These methods excel in detecting anomalies based on temporal patterns but often overlook the structural dependencies between multiple variables, which limits their ability to capture complex, cross-variable relationships. These methods are typically more effective at detecting global anomalies but struggle with structural anomalies involving intricate inter-variable interactions.
\ding{183} \textbf{Graph-based approaches:} In recent years, graph-based methods have become increasingly popular, as they explicitly model the interdependencies between variables using graph structures. Models such as GDN \cite{liu2024multivariate}, Anomaly Transformer \cite{xu2022anomaly-Transformer}, and TranAD \cite{yang2023dcdetector} use graph neural networks (GNNs) and attention mechanisms to model both temporal and spatial dependencies. While these methods are successful in capturing complex inter-variable relationships, most rely on a single-scale graph representation (either static or dynamic), which fails to model the multi-scale dependencies present in the data. Moreover, these methods are highly sensitive to short-term noise, which can lead to a loss of accuracy in detecting structural anomalies.
\ding{184}~\textbf{Contrastive learning methods:} Recent advances in self-supervised and contrastive learning \cite{oord2018cpc,tang2024perturbation} have shown promise in anomaly detection tasks. Methods like OmniAnomaly \cite{xu2024couta} and DCDetector \cite{yang2023dcdetector} use contrastive learning to improve the discriminative ability of the learned representations. However, traditional contrastive learning techniques often operate at the sample level, making them sensitive to short-term perturbations, which can reduce their ability to detect more complex structural anomalies in MTS data.
Subsequent studies also show that such reliance on instance-level semantics can ignore inherent temporal correlations \cite{lee2024softclt}, undermining robust anomaly detection.

Unlike prior work, we learn hierarchical dynamic graphs—local, regional, and global rather than a single static or windowed graph~\cite{deng2021graph,zheng2023correlation,xu2022anomaly-Transformer}; on top of these graphs, we introduce a cross-scale contrastive objective (CDS) that couples intra-scale separation with inter-scale coherence, moving beyond instance-level contrastive learning~\cite{oord2018cpc,yang2023dcdetector} to reveal structure-breaking anomalies; and we align dynamic graphs with EMA-updated stable references (SAA) to suppress noise and drift, unlike methods lacking a structural anchor~\cite{zhou2023mtgflow,dai2022GANF}.

\section{Methodology}\label{sec:method}
\subsection{Problem Definition}\label{sec:problem}
\noindent MTS anomaly detection aims to identify anomalous observations, which may occur either as individual data points or as subsequences that exhibit abnormal temporal or structural patterns in a multivariate system. Formally, let a sliding window be denoted as 
$\mathbf{X}_{1:L} = (\mathbf{x}_1, \mathbf{x}_2, \ldots, \mathbf{x}_L) \in \mathbb{R}^{L \times K}$, 
where $L$ denotes the window length, $K$ is the number of variables, and each 
$\mathbf{x}_t \in \mathbb{R}^K$ represents the $K$-dimensional observation at time step $t$.
The objective is to learn a mapping:
\begin{equation}
f_{\Theta}: \mathbf{X} \mapsto (\mathbf{S}),
\label{eq:mapping-function}
\end{equation}
where $\mathbf{S}$ denotes the anomaly scores at each time step, and $\bm{\Theta}$ is the set of learnable parameters. 
For evaluation purposes, we denote the ground-truth labels as $\mathbf{Y}$, which are not used during training.

\subsection{\texttt{CGSTA} Overview}
\begin{figure}[t]
\centering
\includegraphics[width=\linewidth]{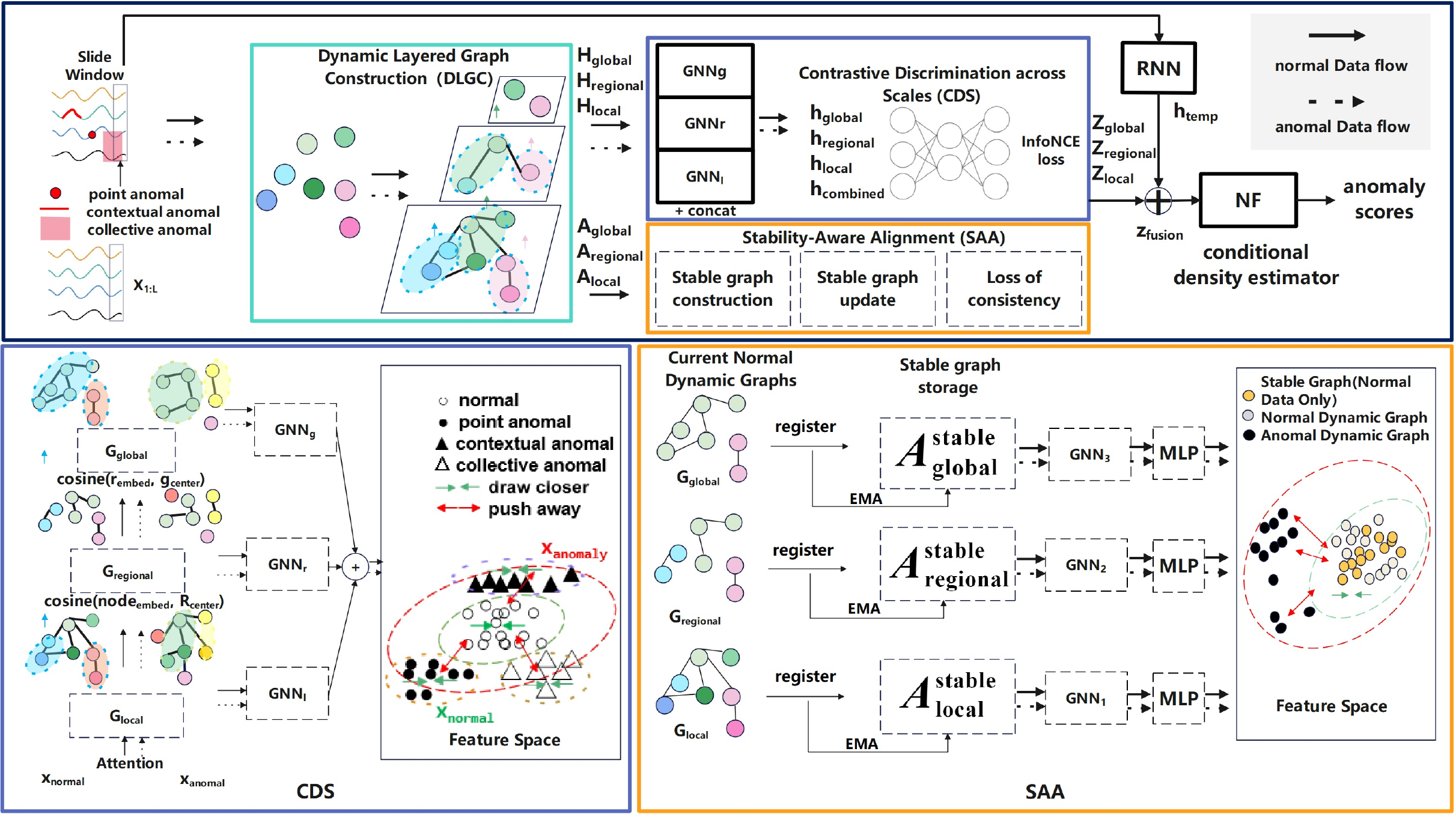}
\caption{Overview of the proposed \texttt{CGSTA} framework for multivariate time series anomaly detection. The framework consists of  (i) DLGC, which builds local, regional, and global graphs with nodes as variables and edges as temporal dependencies; (ii) CDS, which learns discriminative and consistent representations across graph layers; and (iii) SAA maintains stable graphs and enforces consistency with dynamic ones. }
\label{fig:architecture}
\end{figure}
\noindent As shown in Figure~\ref{fig:architecture}, the proposed \texttt{CGSTA} framework integrates three core modules: 
\textbf{Dynamic Layered Graph Construction (DLGC)}, 
\textbf{Contrastive Discrimination across Scales (CDS)}, 
and \textbf{Stability-Aware Alignment (SAA)}. 
At a high level, input sequences are transformed into layered graphs, encoded via GNNs, aligned through contrastive and stability objectives, and scored by a density estimator for anomaly detection. 
This framework explicitly models multivariate time series as graphs, where nodes correspond to variables and edges represent their dependencies. 
By constructing and integrating local, regional, and global graphs, \texttt{CGSTA} captures both short- and long-range correlations and remains robust under dynamically evolving dependency structures.
Prior MTSAD methods often rely on single-scale graph modeling or transformer-style associations, which may miss cross-granularity structures and be sensitive to spurious correlations \cite{deng2021graph,zheng2023correlation,xu2022anomaly-Transformer,liu2024multivariate,Wang2024FCSTGNN}.

\textbf{DLGC:} This module builds hierarchical graphs at three levels—local, regional, and global. Local graphs capture short-term dependencies among variables, while regional and global graphs encode collaborative and system-level patterns via learnable clustering. These graphs are fused with temporal features to yield multi-scale representations for CDS, and their adjacency structures are preserved for SAA.

\textbf{CDS:} This module applies contrastive discrimination on the multi-scale embeddings from DLGC to enhance separability between normal and anomalous patterns. It jointly optimizes intra-scale, inter-scale, and fusion-level objectives, ensuring that anomalies disrupting structural dependencies are explicitly highlighted. The refined embeddings are then fed into the detection head for density-based scoring.

\textbf{SAA:} This module introduces long-term stable graphs, updated via exponential moving average (EMA), as references to the dynamic graphs. By enforcing alignment between dynamic and stable graphs, it reduces sensitivity to noise and prevents anomalies from contaminating structural priors, providing stability-aware supervision for robust representation learning.

\subsection{Dynamic Layered Graph Construction (DLGC)}\label{sec:layered-contrast}
\noindent 
The DLGC module builds hierarchical graphs at three levels (local, regional, and global) to capture dependencies at multiple granularities. Unlike prior works that only model pairwise correlations at a single scale, DLGC explicitly integrates fine-grained, meso-level, and system-wide structures, allowing anomalies of different types to be exposed.

At the local level, we compute a dynamic adjacency matrix from raw input $\mathbf{X} \in \mathbb{R}^{N \times K \times L}$ via attention:
\begin{equation}
\mathbf{A}_{\text{local}} = \text{Attn}(\mathbf{X}),
\end{equation}
which reflects time-varying correlations among variables.

At the regional level, variables are clustered into $R$ functional groups using learnable embeddings and centers:
\begin{equation}
\phi(i) = \arg\max_{r} \big( \mathbf{E}[i] \cdot \mathbf{C}_{\text{regional}}[r] \big),
\end{equation}
where $\phi(i)$ denotes the regional assignment of variable $i$. Two variables that belong to the same region are connected, forming $\mathbf{A}_{\text{regional}}$ (intuitively, regions correspond to functional groups of sensors).  

At the global level, regional embeddings are further aggregated into $G$ clusters:
\begin{equation}
\mathbf{E}_{\text{regional}}[r] = \tfrac{1}{|\mathcal{R}_r|}\sum_{i \in \mathcal{R}_r}\mathbf{E}[i],
\end{equation}
and linked via global centers $\mathbf{C}_{\text{global}}$ to yield $\mathbf{A}_{\text{global}}$ (global clusters represent collections of related regions).  

Finally, graph convolutions are applied on $\mathbf{A}_{\text{local}}$, $\mathbf{A}_{\text{regional}}$, and $\mathbf{A}_{\text{global}}$ to produce multi-scale representations $\mathbf{H}_{\text{local}}$, $\mathbf{H}_{\text{regional}}$, and $\mathbf{H}_{\text{global}}$. These serve as inputs to CDS for contrastive learning, while the adjacency matrices are preserved for SAA to enforce stability.

\subsection{Contrastive Discrimination across Scales (CDS)}\label{sec:multiscale-contrast}
\noindent
Anomalies in MTS may appear inconspicuous at a single scale but become evident when examined through multiple perspectives. Most existing contrastive methods focus on instance-level similarity, but largely ignore hierarchical graph structures that naturally capture such multi-scale dependencies \cite{tang2024perturbation,yang2023dcdetector,qin2023multiview,duan2023graph,darban2025carla}. 
By contrast, CDS leverages local, regional, and global graph embeddings to enhance discriminative ability and cross-scale consistency, and consists of three complementary objectives: intra-scale discrimination, inter-scale consistency, and fusion-level discrimination.We generate abnormal samples based on the original normal data through three data augmentation strategies (point anomaly injection, context replacement, and cluster drift), without involving real anomaly labels.

Our CDS module learns to distinguish normal and anomalous patterns through multi-scale contrastive learning. Its objective comprises three components:

\textbf{Intra-scale discrimination.} 
To learn class-discriminative representations at each scale, we leverage two augmented views of each sample: \textbf{Positive view} (the original normal sample $x_i$) and \textbf{Negative view (pseudo-anomaly)} ($x_i^{\text{aug}}$ generated by applying data augmentations to $x_i$), modeling anomalous patterns without requiring real anomaly labels.

At each scale $\ell \in \{\text{local}, \text{regional}, \text{global}\}$, 
we apply \textbf{symmetric} InfoNCE loss to enforce bidirectional discrimination between normal and pseudo-anomalous views:
\begin{equation}
\mathcal{L}_{\text{intra},\ell} = \frac{1}{2}\Big( \mathcal{L}_{\text{pos}} + \mathcal{L}_{\text{neg}} \Big),
\label{eq:intra-loss}
\end{equation}
where
\begin{equation}
\mathcal{L}_{\text{pos}} = -\frac{1}{N}\sum_{i=1}^{N}
\log \frac{\sum_{k \neq i}^{N} \exp(\text{sim}(z_{\ell,i}^{\text{pos}}, z_{\ell,k}^{\text{pos}})/\tau)}
{\sum_{k \neq i}^{N} \exp(\text{sim}(z_{\ell,i}^{\text{pos}}, z_{\ell,k}^{\text{pos}})/\tau) + \sum_{j=1}^{N} \exp(\text{sim}(z_{\ell,i}^{\text{pos}}, z_{\ell,j}^{\text{neg}})/\tau)},
\end{equation}
and
\begin{equation}
\mathcal{L}_{\text{neg}} = -\frac{1}{N}\sum_{i=1}^{N}
\log \frac{\sum_{k \neq i}^{N} \exp(\text{sim}(z_{\ell,i}^{\text{neg}}, z_{\ell,k}^{\text{neg}})/\tau)}
{\sum_{k \neq i}^{N} \exp(\text{sim}(z_{\ell,i}^{\text{neg}}, z_{\ell,k}^{\text{neg}})/\tau) + \sum_{j=1}^{N} \exp(\text{sim}(z_{\ell,i}^{\text{neg}}, z_{\ell,j}^{\text{pos}})/\tau)}.
\end{equation}

Here, $z_{\ell,i}^{\text{pos}}$ and $z_{\ell,i}^{\text{neg}}$ are embeddings of the positive (normal) and negative (pseudo-anomalous) 
views of sample $i$ at scale $\ell$. The symmetric formulation ensures that: 
(1) normal samples cluster together while separating from pseudo-anomalies, 
and (2) conversely, pseudo-anomalies cluster among themselves while staying distant from normal samples, 
enforcing strong bidirectional separation between the two views.

\textbf{Inter-scale consistency.} 
To ensure learned representations maintain coherence across scales, 
we maximize cosine similarity of embeddings from different scales 
for both normal and pseudo-anomalous samples:
\begin{equation}
\mathcal{L}_{\text{inter}} = \frac{1}{2}\Big(\mathcal{L}_{\text{inter}}^{\text{normal}} + \mathcal{L}_{\text{inter}}^{\text{anom}}\Big),
\end{equation}
where
\begin{equation}
\mathcal{L}_{\text{inter}}^{\text{normal}} = -\frac{1}{3N} \sum_{i=1}^{N} 
\Big[ \cos(z_{\text{local},i}, z_{\text{regional},i}) 
+ \cos(z_{\text{local},i}, z_{\text{global},i}) 
+ \cos(z_{\text{regional},i}, z_{\text{global},i}) \Big],
\end{equation}
and $\mathcal{L}_{\text{inter}}^{\text{anom}}$ is computed similarly on pseudo-anomalous embeddings. 
This symmetric treatment ensures that both normal patterns and their anomalous variants maintain 
multi-scale coherence, preventing the model from developing scale-specific artifacts.

\textbf{Multi-scale representation fusion and fusion-level discrimination.} 
We concatenate embeddings from all three scales at the feature level and apply global average pooling 
over sensor and temporal dimensions to obtain a unified multi-scale representation:
\begin{equation}
\mathbf{h}_{\text{concat}, i} = \text{Concat}(h_{\text{local},i}, h_{\text{regional},i}, h_{\text{global},i}) \in \mathbb{R}^{K \times L \times 3H},
\end{equation}
where $h_{\ell,i} \in \mathbb{R}^{K \times L \times H}$ is the multi-step representation from scale $\ell$. 
The fused embedding is then obtained via:
\begin{equation}
\mathbf{z}_{\text{fusion}, i} = \frac{1}{KL} \sum_{k=1}^{K} \sum_{\ell=1}^{L} 
\mathbf{h}_{\text{concat}, i}[k,\ell,:],
\label{eq:fusion}
\end{equation}
yielding $\mathbf{z}_{\text{fusion}, i} \in \mathbb{R}^{3H}$, where $K$ and $L$ are the numbers of sensors 
and time steps, respectively.

On the fused representation, we apply symmetric InfoNCE loss using the same positive/negative pair construction 
as the intra-scale level, ensuring multi-scale feature coherence and anomaly discrimination.

The overall CDS objective jointly optimizes these three terms:
\begin{equation}
\mathcal{L}_{\text{CDS}} = \sum_{\ell} \mathcal{L}_{\text{intra}, \ell} + \mathcal{L}_{\text{inter}} + \mathcal{L}_{\text{fusion}},
\label{eq:cds-total}
\end{equation}
enabling the model to learn discriminative and coherent multi-scale features for robust self-supervised 
anomaly detection without requiring annotated anomaly labels.

\subsection {Stability-Aware Alignment (SAA)}\label{sec:dynamic-stable}
\noindent
While dynamic graphs $\{\mathbf{A}_{\ell}^{\text{dyn}}\}$ (\(\ell \in \{\text{local}, \text{regional}, \text{global}\}\)) capture short-term and time-varying dependencies, they are highly sensitive to transient noise and fluctuations. 
In contrast, long-term stable dependencies provide reliable structural priors but may fail to promptly reflect recent changes. 
To leverage the strengths of both, we propose the SAA module, which maintains stable graphs via EMA and aligns them with their dynamic counterparts. This design ensures robust representation learning while preventing anomalies or noise from contaminating the underlying structure.

\textbf{Stable graph update via exponential moving average (EMA).}
For each scale $\ell$, we maintain a stable adjacency matrix $\mathbf{A}_{\ell}^{\text{stable}}$ updated from the dynamic one $\mathbf{A}_{\ell}^{\text{dyn}}$ via EMA:
\begin{equation}
\mathbf{A}_{\ell}^{\text{stable}} 
\leftarrow \gamma \,\mathbf{A}_{\ell}^{\text{stable}} 
+ (1-\gamma)\,\mathbf{A}_{\ell}^{\text{dyn}},
\label{eq:ema-update}
\end{equation}
where $\gamma \in [0,1)$ controls the update rate. 
This mechanism accumulates long-term dependency information while smoothing out abrupt noise.

\textbf{Consistency loss on GCN embeddings.}
We enforce consistency between embeddings derived from dynamic and stable graph representations at the feature level. 
Let $\mathbf{H}_{\ell}^{\text{dyn}}$ and $\mathbf{H}_{\ell}^{\text{stable}}$ denote the GCN-encoded features based on $\mathbf{A}_{\ell}^{\text{dyn}}$ and $\mathbf{A}_{\ell}^{\text{stable}}$, respectively. 
We define the consistency loss as:
\begin{equation}
\mathcal{L}_{\text{consist}} 
= -\frac{1}{N}\sum_{i=1}^{N}\sum_{\ell}
\cos\big(\mathbf{h}_{\ell,i}^{\text{dyn}}, \mathbf{h}_{\ell,i}^{\text{stable}}\big),
\label{eq:consist-loss}
\end{equation}
where $\cos(\cdot,\cdot)$ denotes cosine similarity and $\mathbf{h}_{\ell,i}$ is the GCN representation of the $i$-th sample at scale $\ell$. 
This term encourages the learned features to preserve essential dependencies across both dynamic and stable graph views, 
acting as a regularizer to prevent rapid feature oscillations driven by noise.

\textbf{Contrastive loss on graph structures.}
Beyond consistency of embeddings, we enhance learning by explicitly contrasting graph structures themselves. 
We flatten each adjacency matrix into a vector and project it to a contrastive space, then apply InfoNCE loss. 
Positive pairs consist of the dynamic and stable graphs from the same sample: $(\mathbf{g}_{i}^{\text{dyn}}, \mathbf{g}_{i}^{\text{stable}})$, 
where $\mathbf{g}$ denotes the flattened and projected graph representation. 
Negative pairs are constructed from pseudo-anomalous dynamic graphs and the stable reference: $(\mathbf{g}_{i}^{\text{dyn,aug}}, \mathbf{g}^{\text{stable}})$.

We implement this with InfoNCE on the graph-level:
\begin{equation}
\mathcal{L}_{\text{contrast}} 
= -\frac{1}{N}\sum_{i=1}^{N}
\log \frac{\exp(\text{sim}(\mathbf{g}_{i}^{\text{dyn}}, \mathbf{g}_{i}^{\text{stable}})/\tau)}
{\exp(\text{sim}(\mathbf{g}_{i}^{\text{dyn}}, \mathbf{g}_{i}^{\text{stable}})/\tau) + \sum_{j=1}^{M} \exp(\text{sim}(\mathbf{g}_{j}^{\text{dyn,aug}}, \mathbf{g}^{\text{stable}})/\tau)},
\label{eq:contrast-loss}
\end{equation}
where $M$ is the number of pseudo-anomalous samples, and $\tau$ is the temperature parameter.
This formulation pulls the dynamic graph of a normal sample toward its stable counterpart while pushing 
pseudo-anomalous dynamic graphs away from the stable structure, thus amplifying structural deviations caused by anomalies.

\textbf{Total SAA loss.}
The overall loss for SAA combines the consistency (on embeddings) and contrastive (on structures) terms:
\begin{equation}
\mathcal{L}_{\text{SAA}}
= \mathcal{L}_{\text{consist}} + \mathcal{L}_{\text{contrast}},
\label{eq:total-saa-loss}
\end{equation}

Through SAA, local-level stable graphs capture long-term variable dependencies, region-level graphs encode persistent inter-region interactions, and global-level stable graphs characterize system-wide stability. 

During training, the consistency and contrastive losses are computed against the previous stable graphs (with stop-gradient on the stable branch to prevent target leakage), and the EMA update is applied after backpropagation to ensure that the stable graphs represent a frozen long-horizon reference during each training step.

\subsection{Overall Training Objective}\label{sec:objective}
\noindent
The proposed framework integrates three complementary objectives: 
the multi-scale contrastive loss $\mathcal{L}_{\text{CDS}}$ (Sec.~\ref{sec:multiscale-contrast}), 
the dynamic--stable consistency loss $\mathcal{L}_{\text{SAA}}$ (Sec.~\ref{sec:dynamic-stable}), 
and the detection loss $\mathcal{L}_{\text{det}}$ from the conditional density estimator. 
The total training objective is defined as:
\begin{equation}
\mathcal{L}_{\text{total}}
= \mathcal{L}_{\text{det}} + \alpha\,\mathcal{L}_{\text{CDS}} + \beta\,\mathcal{L}_{\text{SAA}},
\label{eq:total-objective}
\end{equation}
where $\alpha$ and $\beta$ are non-negative hyperparameters that balance the contributions of contrastive learning and consistency regularization.

\section{Experiments}\label{sec:exp}
\subsection{Experimental setup}
\noindent\textbf{Datasets.}
We evaluate on four widely used real-world benchmarks that cover IT services, industrial control systems, and space telemetry:
(1) \textbf{PSM} \cite{abdulaal2021practical}: a large-scale production system metrics dataset with multivariate KPIs from online services;
(2) \textbf{SWaT} \cite{goh2016swat}: a secure water treatment plant testbed with rich process and sensor variables;
(3) \textbf{WADI} \cite{adepu2017wadi}: a water distribution testbed extending SWaT with more components and longer attack scenarios;
(4) \textbf{SMAP} \cite{hundman2018detecting}: NASA satellite telemetry with diverse subsystem channels.
For all datasets we follow follow a fixed data split: for each dataset we use 60\%  for training, 20\% for validation and the remaining 20\% for testing.

\textbf{Implementation Details.}
To ensure a like-for-like comparison, we train and evaluate all baselines under a unified pipeline built on the MTGFLOW~\cite{zhou2023mtgflow} dataloader, using identical sliding-window generation, per-variable $z$-score normalization computed from training data, mini-batching, and a common procedure for converting model outputs to point-wise anomaly scores. Where prior papers prescribe dataset-specific preprocessing or post-processing (e.g., window length/stride, normalization domain, score smoothing/thresholding), we adopt those settings when compatible; otherwise, we fall back to the shared defaults. All experiments were conducted on a system running Ubuntu 20.04.6 LTS, equipped with an NVIDIA GeForce RTX 3090 GPU (with CUDA 12.8 and 24 GB of VRAM) and 32 GB of system RAM.

\textbf{Metrics. }
Following common practice in recent MTSAD studies~ \cite{xu2022anomaly-Transformer,zhou2023mtgflow,dai2022GANF,liu2025gcad}, we report AUROC and AUPRC computed on point-wise anomaly scores, together with point-level F1 at an operating threshold. AUROC summarizes the ranking quality of scores without fixing a threshold, while AUPRC is especially informative under the heavy class imbalance typical of anomaly detection because it emphasizes performance on the positive class. Point-level F1 complements these threshold-agnostic metrics by reflecting a concrete precision–recall trade-off that matters in deployment. All results are averaged over 5 random seeds; we report mean~$\pm$~std and conduct paired $t$-tests against the runner-up, marking $p{<}0.05/0.01$ as \ding{52}/\ding{52}\ding{52} in Table~\ref{tab:main-results-multi-blocks} to indicate statistical significance. 

\textbf{Baselines.}
We compare \texttt{CGSTA} with seven representative multivariate time-series anomaly detectors. \textbf{Anomaly Transformer}~\cite{xu2022anomaly-Transformer} models temporal associations via attention and detects anomalies by measuring association discrepancy.
\textbf{SARAD}
~\cite{dai2024sarad} integrates spatial association modeling with a progression-style autoencoding scheme to reconstruct normal patterns.
\textbf{CATCH}~\cite{qiu2025tab－CATCH} enhances discriminative temporal representations through frequency-domain blocking and channel-aware fusion.
\textbf{FCSTGNN}~\cite{Wang2024FCSTGNN} treats variables as a fully connected spatio-temporal graph and applies message passing across time and channels.
\textbf{GCAD}~\cite{liu2025gcad} builds Granger-causality graphs and scores deviations from causality-consistent behaviors.
\textbf{GANF}~\cite{dai2022GANF} augments normalizing flows with graph priors to model the density of normal data.
\textbf{MTGFLOW}~\cite{zhou2023mtgflow} learns dynamic graphs and entity-aware flow transformations to capture distribution shifts over time.

\subsection{Main Results}
Table~\ref{tab:main-results-multi-blocks} summarizes the results on four benchmarks. 
\texttt{CGSTA} consistently achieves the best or highly competitive performance across AUROC, AUPRC, and F1 under identical training and evaluation protocols. 
On PSM and WADI, \texttt{CGSTA} yields statistically significant improvements over the strongest baselines. 
On SWaT and SMAP, our model still achieves the highest mean scores, though the gains are not statistically significant because these datasets exhibit higher intrinsic stochasticity—SWaT contains short and sparse attack windows, while SMAP involves heterogeneous telemetry channels with persistent sensor noise. 
Even under such noisy and highly variable conditions, \texttt{CGSTA} maintains stable rankings and small deviations (e.g., SWaT F1 $42.24{\pm}5.79$, SMAP F1 $40.42{\pm}2.13$), confirming its robustness across seeds. 
These results demonstrate that DLGC and CDS enhance discriminability under evolving dependencies, while the SAA mitigates transient noise and graph drift at the representation level, enabling consistent AUROC, AUPRC, and F1 gains across diverse domains.
\begin{table*}[!htbp]
\centering
\scriptsize
\setlength{\tabcolsep}{4pt}
\caption{Experimental results on four datasets (mean $\pm$ std). Best per metric within each dataset in \textbf{bold}; runner-up is \uline{underlined}. Checkmarks in the last row denote significance vs.\ runner-up: \ding{52} ($p{<}0.05$), \ding{52}\ding{52} ($p{<}0.01$).}
\resizebox{1\linewidth}{!}{
\begin{tabular}{l c c c c c c}
\toprule
\multirow{2}{*}{\textbf{Methods}} & \multicolumn{3}{c}{\textbf{PSM (Pooled Server Metrics)}} & \multicolumn{3}{c}{\textbf{WADI (Water Distribution)}} \\
\cmidrule(lr){2-4} \cmidrule(lr){5-7}
& \textbf{ROC (\%)} & \textbf{PRC (\%)} & \textbf{F1 (\%)} & \textbf{ROC (\%)} & \textbf{PRC (\%)} & \textbf{F1 (\%)} \\
\midrule
Anomaly-Transformer (ICLR'22) & 70.78$\pm$2.70 & 69.99$\pm$4.67 & 62.20$\pm$2.56 & 80.54$\pm$1.38 & 41.47$\pm$2.49 & 43.91$\pm$2.69 \\
GANF (ICLR'22)                & 84.65$\pm$2.17 & 77.52$\pm$4.53 & 72.83$\pm$1.50 & \uline{88.26$\pm$0.72} & \uline{48.40$\pm$2.58} & 56.05$\pm$3.75 \\
MTGFLOW (AAAI'23)             & 84.21$\pm$2.78 & 79.88$\pm$3.66 & 71.93$\pm$1.17 & 85.98$\pm$1.90 & 45.70$\pm$1.73 & \uline{56.43$\pm$2.10} \\
FCSTGNN (AAAI'24)             & 64.77$\pm$3.32 & 58.56$\pm$5.74 & 57.67$\pm$2.20 & 79.72$\pm$12.68 & 44.01$\pm$17.44 & 46.35$\pm$16.17 \\
SARAD (NeurIPS'24)            & \uline{85.95$\pm$1.89} & \uline{82.49$\pm$2.04} & \uline{75.28$\pm$1.68} & 71.03$\pm$2.33 & 21.47$\pm$0.88 & 36.53$\pm$1.57 \\
CATCH (ICLR'25)               & 71.31$\pm$0.23 & 67.09$\pm$0.19 & 61.23$\pm$0.16 & 58.69$\pm$0.35 & 16.59$\pm$0.15 & 27.95$\pm$0.45 \\
GCAD (AAAI'25)                & 78.01$\pm$2.42 & 65.77$\pm$4.63 & 67.81$\pm$2.33 & 83.42$\pm$0.72 & 35.52$\pm$2.56 & 41.36$\pm$3.34 \\
\textbf{\texttt{CGSTA} (Ours)} & \textbf{87.93$\pm$1.43} & \textbf{83.03$\pm$2.63} & \textbf{76.24$\pm$1.87} & \textbf{89.67$\pm$1.34} & \textbf{55.87$\pm$7.26} & \textbf{59.17$\pm$6.30} \\
\textbf{Statistical Significance} & \ding{52}\ding{52} & \ding{52} & \ding{52}\ding{52} & \ding{52} & \ding{52} & \ding{52} \\
\midrule
\multirow{2}{*}{\textbf{Methods}} & \multicolumn{3}{c}{\textbf{SWaT (Secure Water Treatment)}} & \multicolumn{3}{c}{\textbf{SMAP (Soil Moisture Active Passive)}} \\
\cmidrule(lr){2-4} \cmidrule(lr){5-7}
& \textbf{ROC (\%)} & \textbf{PRC (\%)} & \textbf{F1 (\%)} & \textbf{ROC (\%)} & \textbf{PRC (\%)} & \textbf{F1 (\%)} \\
\midrule
Anomaly-Transformer (ICLR'22) & 80.79$\pm$5.17 & 33.32$\pm$12.33 & 36.33$\pm$9.65 & 49.39$\pm$0.93 & 14.98$\pm$0.35 & 31.89$\pm$1.39 \\
GANF (ICLR'22)                & 79.74$\pm$2.53 & 22.35$\pm$3.58 & 32.16$\pm$3.43 & 52.04$\pm$4.71 & 15.74$\pm$2.31 & 28.75$\pm$1.33 \\
MTGFLOW (AAAI'23)             & \uline{83.90$\pm$0.79} & \uline{34.58$\pm$3.62} & \uline{41.86$\pm$2.68} & 62.15$\pm$5.64 & 21.85$\pm$5.43 & 34.84$\pm$5.77 \\
FCSTGNN (AAAI'24)             & 68.93$\pm$3.57 & 18.98$\pm$6.60 & 24.44$\pm$5.31 & 62.94$\pm$5.42 & 20.51$\pm$4.16 & 34.53$\pm$3.20 \\
SARAD (NeurIPS'24)            & 71.79$\pm$0.63 & 11.27$\pm$0.42 & 21.27$\pm$1.66 & 51.33$\pm$8.78 & 15.50$\pm$2.43 & 30.86$\pm$3.02 \\
CATCH (ICLR'25)               & 66.59$\pm$0.15 & 9.26$\pm$0.03 & 17.71$\pm$0.07 & 56.41$\pm$0.06 & 16.41$\pm$0.02 & 33.51$\pm$0.00 \\
GCAD (AAAI'25)                & 73.32$\pm$3.98 & 23.10$\pm$3.73 & 32.16$\pm$4.25 & \uline{68.81$\pm$1.80} & \uline{27.45$\pm$4.22} & \uline{39.66$\pm$4.65} \\
\textbf{\texttt{CGSTA} (Ours)} & \textbf{84.77$\pm$1.76} & \textbf{37.58$\pm$7.07} & \textbf{42.24$\pm$5.79} & \textbf{70.20$\pm$2.25} & \textbf{28.65$\pm$6.86} & \textbf{40.42$\pm$2.13} \\
\textbf{Statistical Significance} & -- & -- & -- & -- & -- & -- \\
\bottomrule
\end{tabular}
}
\label{tab:main-results-multi-blocks}
\end{table*}
\FloatBarrier
\subsection{Ablation study}\label{subsec:ablation}
\textbf{Variants.}
We evaluate four configurations to quantify the contribution of each module:Full (DLGC + CDS + SAA): Uses the complete hierarchy, cross-scale contrast, and stability-aware alignment. \textit{w/o} SAA: Removes the stability-aware branch; the EMA updates, dynamic–stable consistency, and dynamic–stable contrastive losses are disabled, while DLGC and CDS remain unchanged.{\textit{w/o} CDS: Drops all cross-scale contrastive terms (intra-, inter-, and fusion-level); the model is trained only with the detection loss plus DLGC.
\textbf{single-scale only}: Trained solely with the detection loss.

\textbf{Findings.}
Tables~\ref{tab:ablation-3mods-psm-wadi} and~\ref{tab:ablation-3mods-swat-smap} show that removing any major module consistently degrades performance across all datasets, indicating complementary effects. 
Removing any major component (DLGC, CDS, or SAA) consistently leads to a decrease in performance across all datasets, highlighting the complementary nature of the modules. Removing SAA leads to a notable reduction in performance, especially in terms of handling transient noise and drift. This supports its role in stabilizing the learned dependencies by maintaining a reference graph. Eliminating CDS, which is responsible for cross-scale contrast, results in a weakened ability to differentiate between normal and anomalous patterns. This emphasizes the importance of cross-scale alignment in enhancing structure-aware learning. Simplifying the graph structure by removing DLGC, which builds multi-scale views of the data, leads to the most significant performance degradation. This illustrates the necessity of multi-scale information, particularly in datasets with complex inter-variable dependencies. 
\begin{table}[htb]
\centering
\scriptsize
\setlength{\tabcolsep}{5pt}
\caption{Ablation on the three main modules (PSM  \& WADI). Mean$\pm$std over 5 seeds. Best in \textbf{bold}, runner-up \uline{underlined}. “\wout” denotes removing the module.}
\begin{tabular}{lcccccc}
\toprule
\multirow{2}{*}{\textbf{Variants}} & \multicolumn{3}{c}{\textbf{PSM (Pooled Server Metrics)}} & \multicolumn{3}{c}{\textbf{WADI (Water Distribution)}} \\
\cmidrule(lr){2-4}\cmidrule(lr){5-7}
& \textbf{ROC (\%)} & \textbf{PRC (\%)} & \textbf{F1 (\%)} & \textbf{ROC (\%)} & \textbf{PRC (\%)} & \textbf{F1 (\%)} \\
\midrule
\textbf{Full (DLGC + CDS + SAA)} 
& \textbf{87.93$\pm$1.43} & \textbf{83.03$\pm$2.63} & \textbf{76.24$\pm$1.87}
& \textbf{89.67$\pm$1.34} & \textbf{55.87$\pm$7.26} & \textbf{59.17$\pm$6.30} \\
\quad \wout~SAA                      
& \uline{87.01$\pm$1.46} & \uline{81.38$\pm$2.39} & \uline{74.55$\pm$1.46}
& \uline{88.74$\pm$1.00} & \uline{49.69$\pm$4.56} & \uline{56.59$\pm$5.30} \\
\quad \wout~CDS                      
& 85.69$\pm$1.37 & 78.85$\pm$3.06 & 73.16$\pm$1.10
& 88.38$\pm$1.26 & 48.07$\pm$2.10 & 56.39$\pm$4.43 \\
\quad \wout~DLGC (single-scale only) 
& 85.00$\pm$1.43 & 78.25$\pm$2.77 & 72.47$\pm$1.35
& 88.43$\pm$1.82 & 49.58$\pm$5.32 & 56.33$\pm$5.00 \\
\bottomrule
\end{tabular}
\label{tab:ablation-3mods-psm-wadi}
\end{table}
\begin{table}[htb]
\centering
\scriptsize
\setlength{\tabcolsep}{5pt}
\caption{Ablation on the three main modules (SWaT \& SMAP). Mean$\pm$std over 5 seeds. “\wout” denotes removing the module.}
\begin{tabular}{lcccccc}
\toprule
\multirow{2}{*}{\textbf{Variants}} & \multicolumn{3}{c}{\textbf{SWaT (Secure Water Treatment)}} & \multicolumn{3}{c}{\textbf{SMAP (Soil Moisture Active Passive)}} \\
\cmidrule(lr){2-4}\cmidrule(lr){5-7}
& \textbf{ROC (\%)} & \textbf{PRC (\%)} & \textbf{F1 (\%)} & \textbf{ROC (\%)} & \textbf{PRC (\%)} & \textbf{F1 (\%)} \\
\midrule
\textbf{Full (DLGC + CDS + SAA)} 
& \textbf{84.77$\pm$1.76} & \textbf{37.58$\pm$7.07} & \textbf{42.24$\pm$5.79}
& \textbf{70.20$\pm$2.25} & \textbf{28.65$\pm$6.86} & \textbf{40.42$\pm$2.13} \\
\quad \wout~SAA                      
& \uline{84.47$\pm$2.01} & \uline{37.15$\pm$3.15} & \uline{41.73$\pm$2.40}
& \uline{68.21$\pm$2.53} & 26.35$\pm$4.84 & \uline{39.22$\pm$0.82} \\
\quad \wout~CDS                      
& 81.36$\pm$1.35 & 31.65$\pm$8.15 & 39.89$\pm$5.61
& 67.52$\pm$1.20 & \uline{28.28$\pm$5.79} & 38.29$\pm$3.09 \\
\quad \wout~DLGC (single-scale only) 
& 81.32$\pm$2.33 & 30.34$\pm$4.84 & 36.73$\pm$2.29
& 63.57$\pm$3.76 & 23.59$\pm$5.01 & 34.63$\pm$2.77 \\
\bottomrule
\end{tabular}
\label{tab:ablation-3mods-swat-smap}
\end{table}

\subsection{Hyperparameter Sensitivity}
\textbf{Sensitivity to CDS weight $\alpha$.}
We sweep $\alpha\in\{0.70,0.75,0.80,0.85,0.90,0.95\}$ on PSM, WADI, SWaT, and SMAP
and report AUROC and AUPRC (mean$\pm$std over five runs). As shown in
Figure~\ref{fig:alpha_sensitivity}, the curves are generally flat, indicating that
the model is not overly sensitive to $\alpha$. PSM and WADI show a mild upward
trend in AUPRC as $\alpha$ approaches $0.90$; SWaT remains largely unchanged
across the range; SMAP varies slightly with a small improvement at the higher
end. Overall, a mid-range $\alpha$ delivers robust performance across datasets.
\begin{figure}[htbp!]
  \centering
  \includegraphics[width=0.9\linewidth]{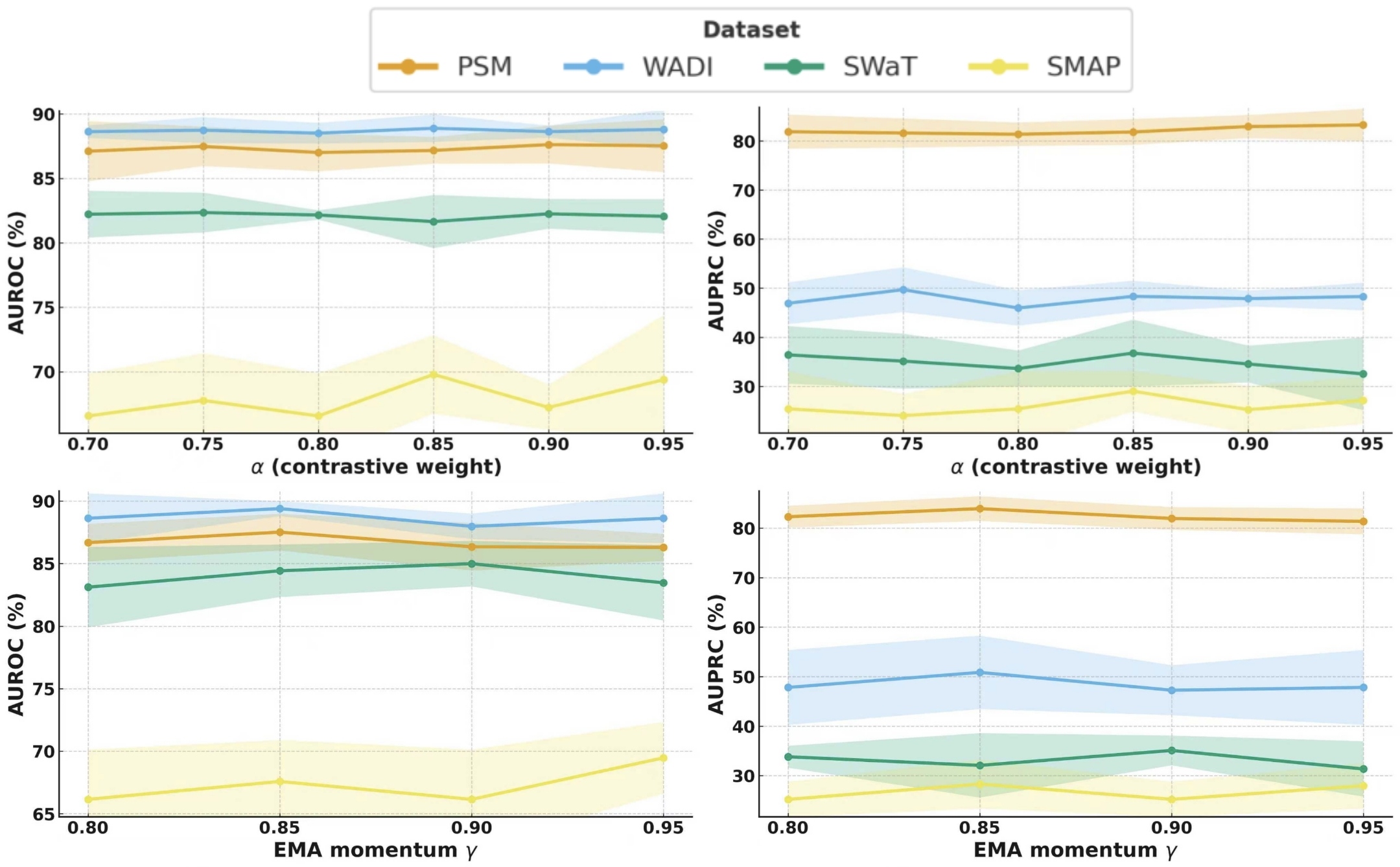}
  \caption{Sensitivity of AUROC and AUPRC to Hyperparameters: CDS weight $\alpha$ ($[0.70,0.95]$) and EMA momentum $\gamma$ ($[0.80,0.95]$). Results are shown as mean$\pm$std across PSM, WADI, SWaT, and SMAP (left: AUROC; right: AUPRC).}
  \label{fig:alpha_sensitivity}
\end{figure}
\textbf{EMA Momentum Sensitivity.}
The EMA momentum $\gamma\in[0,1)$ controls the update rate of the stable graphs. We sweep $\gamma\in{0.80,0.85,0.90,0.95}$ on PSM, WADI, SWaT, and SMAP and report AUROC and AUPRC (mean±std over seeds). As shown in Figure~\ref{fig:alpha_sensitivity}, performance varies smoothly with $\gamma$. A moderate momentum ($\gamma \approx 0.85$) is a robust default on PSM and WADI (highest averages across metrics), SWaT peaks around $\gamma = 0.90$ on AUROC/AUPRC, while SMAP benefits from stronger smoothing with best AUROC at $\gamma = 0.95$. These results indicate that mild-to-moderate EMA smoothing stabilizes the dynamic graphs without over-damping dataset-specific patterns.

\textbf{Sensitivity of SAA's Weight.}
The SAA's loss coefficient $\beta$ controls the strength of aligning dynamic graphs to the stable graph.
Because the feasible ranges differ notably across datasets. Each curve shows the mean across seeds and the shaded band indicates $\pm$ standard deviation. As shown in
Figure~\ref{fig:beta_roc_prc}, performance varies smoothly with~$\beta$:
PSM has a mild peak near $\beta{=}0.35$; WADI benefits from stronger alignment around
$\beta{\approx}0.39$; SWaT remains stable within a low-$\beta$ band; SMAP is shallowly U-shaped with
competitive AUROC near $\beta{\in}\{0.11,0.15\}$. Overall, a dataset-specific, moderate~$\beta$
balances stability and responsiveness, damping noisy graph fluctuations without oversmoothing.
\FloatBarrier
\begin{figure}[htbp!]
  \centering
  \includegraphics[width=0.9\linewidth]{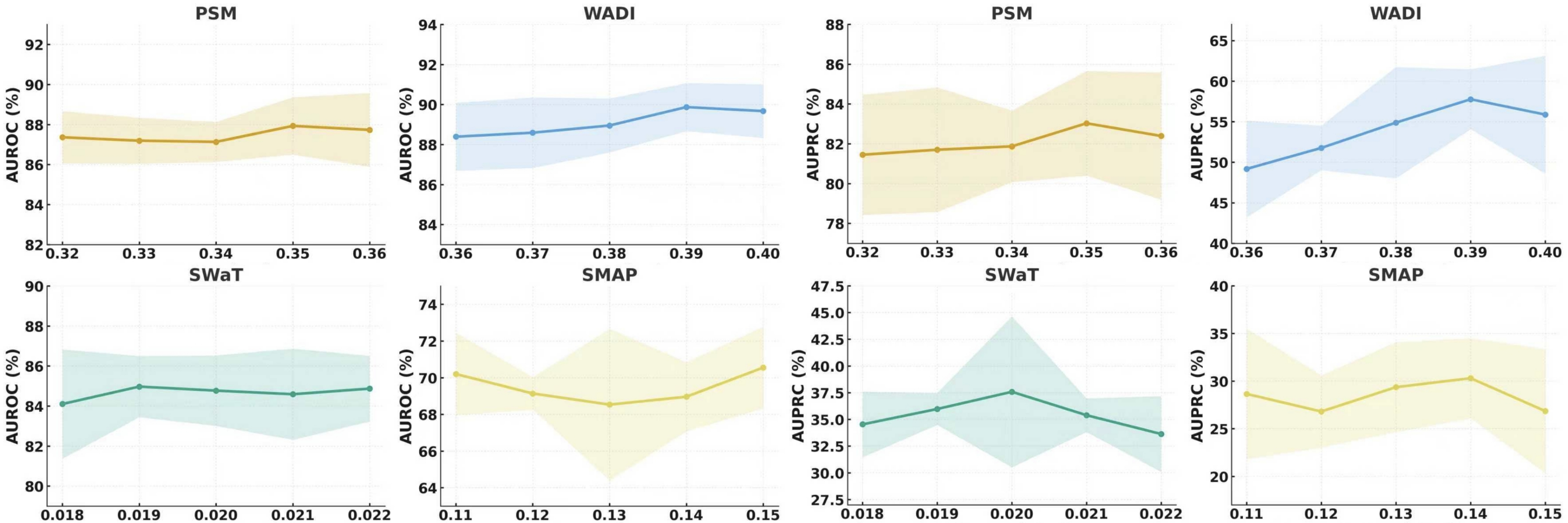}
  \caption{Sensitivity to SAA’s weight $\beta$ — AUROC and AUPRC across datasets}
  \label{fig:beta_roc_prc}
\end{figure}
\FloatBarrier

\subsection{Case Study}
To intuitively demonstrate the role of \texttt{CGSTA} in multi-scale structure modeling and steady-state alignment, we selected a representative attack window from the SWaT test set (sample index 7894, label Attack). We focuses on three points: where the anomaly occurs , how the dependency structure behaves across the local–regional–global levels within this window and whether these changes manifest as structural deviations rather than short-term noise. Figure~\ref{fig:case} summarizes the visualization results.
\begin{figure}[t]
  \centering
  \includegraphics[width=\linewidth]{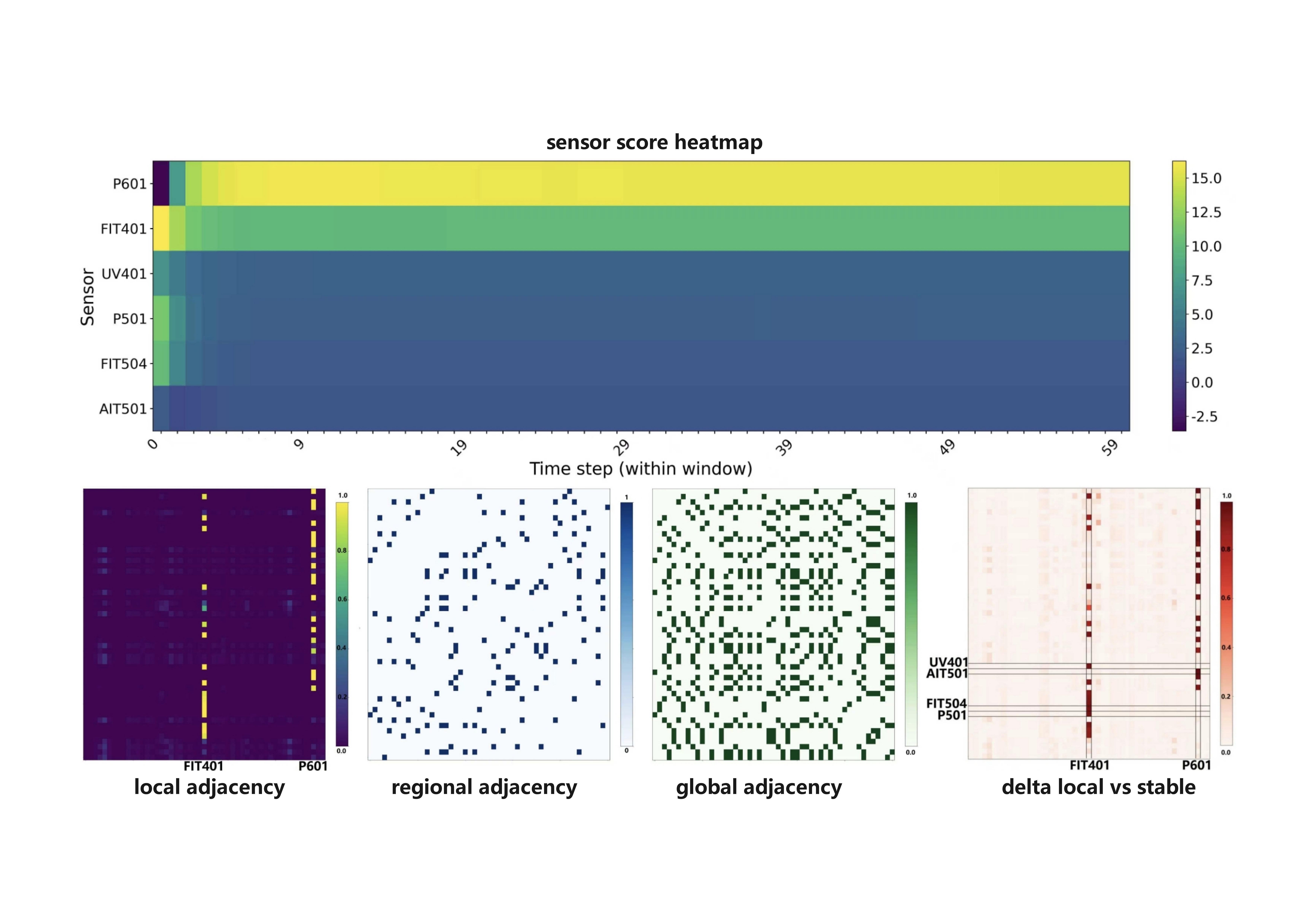}
  \caption{Case on SWaT (index 7894, label Attack).
  Top: per-sensor anomaly scores for Top-$K$ sensors (brighter is more anomalous).
  Bottom (from left to right): dynamic local dependency (row-wise softmax), 
  regional adjacency (group-wise), global adjacency (cross-group), 
  $\Delta_{\text{local}}=|A_{\text{local}}^{\text{dyn}}-A_{\text{local}}^{\text{stable}}|$,
  and the EMA-anchored stable local graph. 
  Bright vertical bands in the dynamic graph indicate hub targets (e.g., P601/FIT401); 
  $\Delta_{\text{local}}$ concentrates on edges incident to these hubs, 
  evidencing structural deviations rather than transient noise.}
  \label{fig:case}
\end{figure}

We visualize a representative attack window (SWaT, index 7894) in Figure~\ref{fig:case} to illustrate \texttt{CGSTA}'s ability to detect structural anomalies.

\textbf{Anomaly localization.}
The top heatmap shows sustained high scores on P601, FIT401, and related sensors, indicating a persistent abnormal regime rather than an isolated spike.

\textbf{Multi-scale dependency (DLGC).}
The dynamic local graph (bottom left) shows bright vertical bands on P601/FIT401, meaning most sensors converge their dependencies to these hubs during the attack. Regional and global adjacencies form block-wise links, reflecting functional-group and system-level couplings around affected modules.

\textbf{Stability-aware detection (SAA).}
The stable graph at each scale is learned by EMA accumulation over
normal-only training batches, encoding the long-term structural
baseline.  During the attack, $\Delta_{\text{local}}$ peaks sharply on
edges connected to P601/FIT401, confirming the anomaly breaks
fine-grained pairwise dependencies rather than producing transient noise. 

\section{Conclusion}
We present \texttt{CGSTA}, an MTS anomaly detector that models multi-scale dependency structures and anchors them to stable priors. Its core modules work in tandem: DLGC builds hierarchical (local-regional-global) graphs per window; CDS enforces structure-aware learning via cross-scale contrast; SAA suppresses noise/drift by aligning dynamic graphs to EMA-stabilized references. A density head then outputs fine-grained anomaly scores.
Across four benchmarks, \texttt{CGSTA} outperforms baselines on structure-rich datasets (PSM, WADI) and remains competitive on sparse/noisy ones (SWaT, SMAP)—suggesting multi-scale graph contrast thrives where structural patterns are clear, but is tempered by extreme sparsity/heterogeneity. Ablations confirm the complementarity of DLGC, CDS, and SAA; our case study further illustrates how \texttt{CGSTA} targets structural deviations.
For future work, we will make hierarchical assignments fully dynamic, optimize streaming efficiency, and boost robustness to data contamination and distribution shifts.

\subsubsection*{Acknowledgments.}
\small
This work was supported by the Fundamental Research Funds of Dalian Maritime University (Grant No. 3132024630). Jun Zhang is the corresponding author.
\subsubsection*{Disclosure of Interests.}
\small
The authors declare no competing interests relevant to the content of this article.

\FloatBarrier
\bibliographystyle{splncs04}

\end{document}